\documentclass{article}

\usepackage{arxiv}
\usepackage[hyphens]{url}
\usepackage[utf8]{inputenc} % allow utf-8 input
\usepackage[T1]{fontenc}    % use 8-bit T1 fonts
\usepackage{hyperref}       % hyperlinks
\usepackage{booktabs}       % professional-quality tables
\usepackage{amsmath}        % math environments
\usepackage{amssymb}        % blackboard math symbols
\usepackage{amsfonts}       % blackboard math symbols
\usepackage{nicefrac}       % compact symbols for 1/2, etc.
\usepackage{microtype}      % microtypography
\usepackage{cleveref}       % smart cross-referencing
\usepackage{graphicx}
\usepackage{natbib}
\usepackage{doi}

\title{Contextual Quality-Diversity Evolutionary Reinforcement Learning for HVAC Control in Tropical Commercial Buildings}

\newif\ifuniqueAffiliation
\uniqueAffiliationtrue

\ifuniqueAffiliation
\author{ Tran Le Vu \\
	Energy Research Institute @ Nanyang Technological University\\
	\#06-04, Cleantech One, 1 Cleantech Loop\\
	Singapore 637141 \\
	\texttt{levu.tran@ntu.edu.sg} \\
}
\fi

\renewcommand{\shorttitle}{CQD-ERL for Tropical HVAC Control}

\hypersetup{
pdftitle={Contextual Quality-Diversity Evolutionary Reinforcement Learning for HVAC Control in Tropical Commercial Buildings},
pdfsubject={eess.SY, cs.LG},
pdfauthor={Tran Le Vu},
pdfkeywords={quality-diversity optimization, evolutionary reinforcement learning, soft actor-critic, HVAC control, chiller plant, tropical buildings, MAP-Elites},
}

\begin{document}
\maketitle

\begin{abstract}
This paper proposes a contextual quality-diversity evolutionary reinforcement-learning controller, CQD-ERL, for the supervisory control of a tropical, water-cooled chiller plant and its associated air side. Rather than converging to a single scalarised policy, the controller maintains a product archive of specialised policies indexed jointly by a data-driven operating context, a cluster of daily weather and load regime, and a context-invariant behaviour descriptor, filled by a gradient-free evolutionary operator and a soft-actor-critic policy-gradient operator that share one replay buffer. Every action is filtered through a deterministic safety shield before execution. The controller is trained on a two-tier reduced-order environment representing the latent load, cooling-tower approach and humidity constraints of a Singapore commercial building, and is evaluated over a full annual backtest against an ASHRAE Guideline 36 baseline. Implementation code is available at \url{https://github.com/stevietran/hvac_cdqerl.git}.
\end{abstract}


\section{Introduction}

Buildings account for a large share of global energy consumption, and within them heating, ventilation and air-conditioning is the dominant end use. The share rises sharply toward the equator. In tropical commercial buildings, air-conditioning can represent 50 to 60\% of total energy, and the central chilled-water plant alone accounts for more than 40\% of space-conditioning energy. A percentage point of efficiency recovered at the plant level is therefore multiplied across an enormous installed base of towers, offices and institutional buildings.

The character of the control problem changes in the humid tropics, in ways a literature written mostly from temperate case studies tends to miss. Three facts dominate. The ambient wet-bulb temperature in Singapore sits near 25 to 26\textdegree C and varies little across the year, so cooling towers reject heat against a persistently narrow and expensive approach, and air-side economising, the dominant temperate energy strategy, contributes almost nothing. The sensible heat ratio of the design load falls to roughly 0.65 to 0.75, so dehumidification rather than temperature sets the coil size and the chilled-water supply temperature. The relative-humidity ceiling near 60 to 65\%, codified locally in SS 553, is a condensation and mould limit rather than a comfort preference, which makes it a hard safety constraint on the controller rather than a soft term to be traded against energy. Benchmark emulators and most published case studies remain overwhelmingly temperate, heating-season and sensible-load dominated, so this regime is systematically under-represented in the evidence base available to a designer.

The incumbent supervisory baseline is a layered set of heuristics, chiller staging on load thresholds, static setpoint optimisation against equipment curves, and the trim-and-respond and dual-maximum sequences of ASHRAE Guideline 36. These sequences are vendor-neutral, field-tested and difficult to beat honestly, and reported machine-learning savings measured against anything weaker are not credible. Model predictive control is the rigorous alternative, and recent field deployments report savings from roughly 11 to 15\% on real chiller plants. Its rigour is nonetheless constrained by structural facts rather than by engineering effort. Every primary study confronts the same structural fact that chiller-plant supervision mixes discrete staging with continuous setpoints through bilinear equipment physics, and an intrinsically slow mixed-integer nonlinear program. Every field deployment approximates this mixed-integer nonlinear program in different ways. The nonlinear-program relaxation dispatched chillers, pumps and thermal storage at an airport central plant, achieving roughly 10\% simulated and 7\% field-verified savings \citep{chinde_model_2024}. Truncating the search to a two-step dynamic program over neural chiller models and gradient-boosted load forecasts achieved 14.98\% field-verified savings while cutting worst-case computation from 10,000 to 2.79 seconds \citep{li_field_2025}. Changing the control signal itself, predicting chilled-water return temperature rather than reacting to it once a limit is crossed, delivered 11.07\% field-verified savings using only sensors already installed \citep{khunmaturod_model_2026}. Learning the mixed-integer policy offline through differentiable predictive control moved the optimisation off the critical path entirely, reaching up to 13\% savings at orders-of-magnitude faster inference than mixed-integer model predictive control, though its plant model omits the cooling tower and condenser loop that bind hardest in a tropical climate \citep{boldocky_data_2026}.

Reinforcement learning removes the requirement for an explicit, continuously maintained plant model, and the strongest results are genuinely encouraging. Applied to buildings, the literature has moved through three phases. Early work demonstrated that model-free deep reinforcement learning could hold a building within a comfort band using weather and price signals alone, reporting energy savings of 20 to 70\% against an unspecified rule-based baseline \citep{wei_deep_2017}. Model-based variants later closed much of the sample-efficiency gap, an ensemble of environment-conditioned dynamics models combined with model-predictive path-integral control matched model-free savings using an order of magnitude less training data \citep{ding_multi-zone_2023}. The most recent phase has tightened the standard of evidence considerably. A physics-informed offline controller ran a production data centre's cooling for more than 2,000 hours and delivered 14 to 21\% savings with no safety violation \citep{zhan_data_2025}, and deep reinforcement learning acting directly on air- and water-side actuators beat Guideline 36 by up to 17\% while generalising to unseen occupancy \citep{savino_deploying_2025}. Yet the field's own audit found that only 11\% of 77 surveyed controllers were ever tested in a real building, tracing this gap to four recurring failure modes: unacceptable exploration cost in an occupied building, non-smooth dynamics such as discrete chiller staging and dew-point cliffs that defeat gradient methods, a single scalarised-reward policy that must compromise across every operating regime a tropical building actually cycles through, and the safety and transfer concerns that block deployment \citep{wang_reinforcement_2020}.

BOPTEST provides a shared infrastructure layer that only recently converged on common standards which supplies containerised Modelica emulators with embedded baseline control, a REST API and fixed key-performance indicators \citep{blum_building_2021}. It finally allows controllers to be benchmarked against each other rather than on bespoke case studies. CityLearn v2 trades physical fidelity for a fast, self-contained, data-driven Gymnasium environment supporting multi-agent and grid-interactive control of building fleets \citep{nweye_citylearn_2025}, and Spawn couples EnergyPlus envelope modelling with Modelica HVAC and control at larger, more tractable coupling time steps than earlier interfaces, so realistic control sequences can be simulated as written rather than approximated \citep{wetter_spawn_2024}. Underlying all three is the question of how much building detail a control-oriented model actually needs. Physics-informed neural networks recover room temperature and hidden thermal-mass state with sub-0.25-degree error and remain accurate at longer horizons than plain networks \citep{gokhale_physics_2022}, and staged grey-box identification shows that a well-identified single-zone model can beat a poorly identified multi-zone one inside model predictive control, so inter-zone detail helps only when it is estimated well \citep{arroyo_identification_2020}.

Outside the building-control literature, a separate line of research has matured a rigorous account of why evolutionary search and gradient reinforcement learning have complementary failure modes, and how the two may be combined. Evolutionary reinforcement learning combines a population-based evolutionary search, gradient-free, globally exploring and indifferent to reward discontinuities, with an off-policy gradient learner that supplies sample efficiency and fine per-step control \citep{khadka_evolution-guided_2018}. Evolution-guided policy gradient lets a population and an off-policy learner share one experience stream, so episode-level credit assignment and broad exploration coexist with gradient sample efficiency, and a portfolio variant reframes hyperparameter sensitivity as a compute-allocation problem across learners of different time horizons \citep{khadka_collaborative_2019}. Quality-diversity search extends this idea by maintaining an archive of diverse, high-performing solutions rather than converging to one policy, using efficient variation operators \citep{vassiliades_discovering_2018} and policy-gradient-assisted variation to scale to deep neural controllers \citep{nilsson_policy_2021,grillotti_quality-diversity_2024}. Within the quality-diversity family, a directed variation operator exploiting the empirical concentration of elites in a shared genotypic region accelerates MAP-Elites substantially \citep{vassiliades_discovering_2018}. A policy-gradient variation operator driven by an asynchronously trained critic scales the archive from roughly one hundred controller parameters to deep networks exceeding twenty thousand \citep{nilsson_policy_2021}. Noise-robust variants instead store a small sub-population per niche rather than a single elite, keeping the archive accurate under noisy fitness at no extra evaluation cost \citep{flageat_fast_2020}. Folding the diversity objective into a single actor-critic through a successor-features critic and a learned constraint multiplier reaches comparable diversity at deep-reinforcement-learning sample efficiency \citep{grillotti_quality-diversity_2024}. A multi-task variant of MAP-Elites replaces the behavioural archive with one indexed by task, solving thousands of related tasks by exploiting the same elite-hypervolume structure the variation operator above assumes \citep{mouret_quality_2020}, later generalised to a continuous task space \citep{anne_parametric-task_2024}. Two recent surveys now organise the whole hybrid field, one by the direction in which evolution and reinforcement learning assist each other \citep{li_bridging_2025} and one by the architectural role evolution plays inside forty-five post-2017 combinations \citep{sigaud_combining_2022}, and a hardware-accelerated library has removed compute as the practical barrier to running any of it \citep{chalumeau_qdax_2023}. Every domain in this lineage remains simulated continuous-control locomotion or manipulation, and its behavioural or task axes have no established building-energy analogue, which is precisely the gap our paper is designed to fill.

This paper proposes a contextual, quality-diversity evolutionary reinforcement-learning controller, CQD-ERL, for a tropical, water-cooled chiller plant and its associated air side, trained and evaluated on a purpose-built reduced-order physics environment calibrated toward Singapore conditions. The contextual concept means the archive is organised around a portfolio of controllers, each specialised to an operating context, and dispatched by matching the current regime to the nearest stored specialist. Soft actor-critic, a deep reinforcement learning algorithm that augments the return with policy entropy, is adopted here \citep{haarnoja_soft_2018}. This off-policy formulation is markedly more stable across random seeds than the deterministic methods underlying most recent HVAC controllers. The environment is designed to be two-tier. A millisecond-cost reduced-order model carries the millions of rollouts population-based search requires, and a calibrated, community-benchmarked twin carries validation, following the BOPTEST protocol. The contribution is threefold: an archive architecture that validates a portfolio of regime-specialised controllers instead of converging to a single compromise policy; a training environment that represents the latent load, condensation risk and cooling-tower approach that dominate tropical operation and that most benchmark emulators omit; and an empirical account, measured against the ASHRAE Guideline 36 baseline.

\section{Contextual Quality-Diversity Evolutionary Reinforcement Learning}
\label{sec:method}

This work uses \emph{contextual} to mean that the deployed controller is not a single policy compromising across every operating condition a tropical building cycles through, but a portfolio of specialised policies selected by the prevailing regime. Context is a low-dimensional summary of the day's weather and load, and it conditions which stored controller is dispatched, rather than being fed as an additional input to one network. The context can be extended to other operating modes such as demand response, electricity tariff structures, and peak shaving.

\begin{figure}[!ht]
	\centering
	\includegraphics[width=\textwidth]{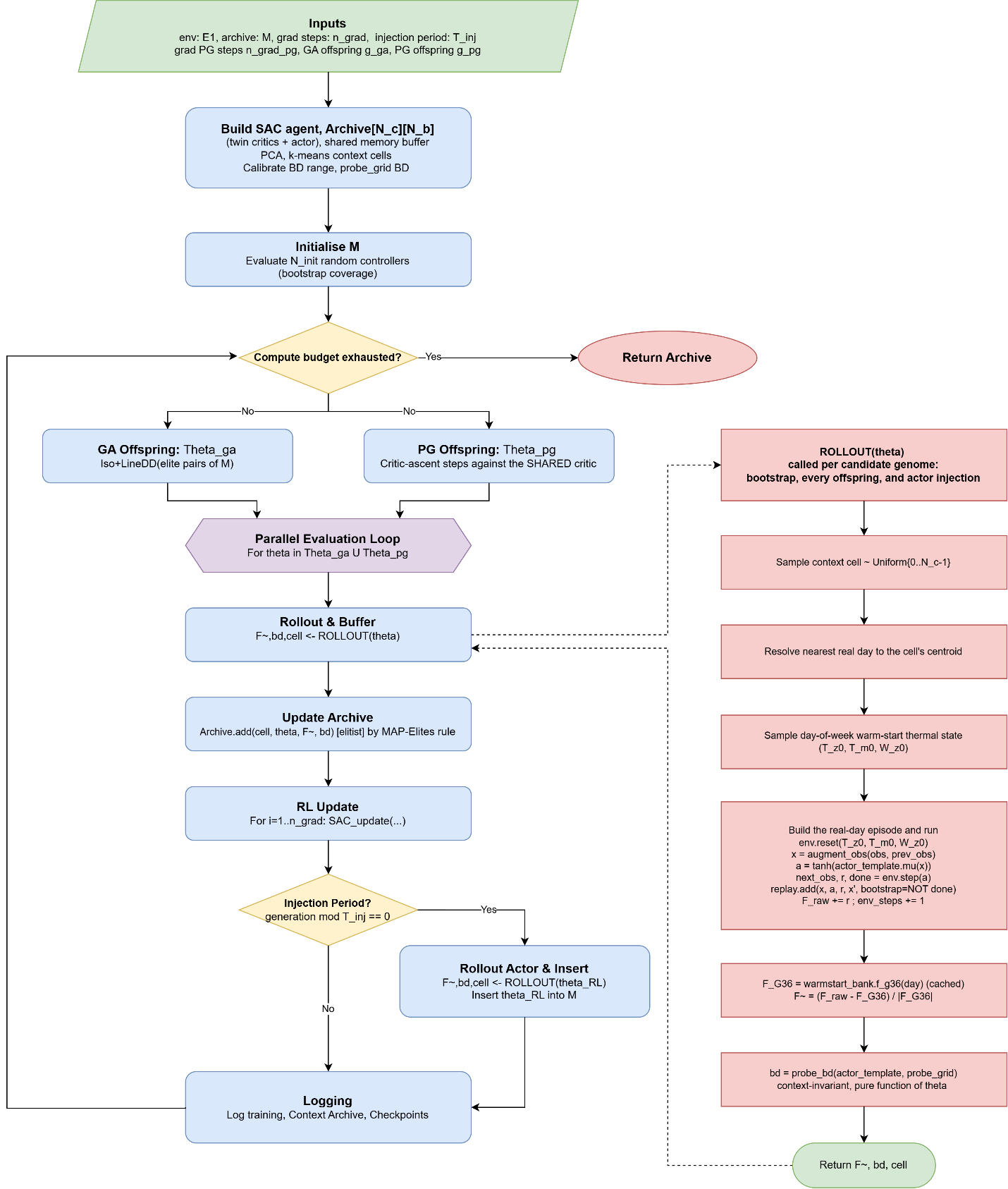}
	\caption{CQD-ERL algorithm diagram.}
	\label{fig:algorithm}
\end{figure}

The controller, termed CQD-ERL, occupies the synergistic direction of the hybrid evolution-reinforcement-learning taxonomy. In this direction, an evolutionary search and a gradient learner both stand alone yet continually exchange information, rather than one merely assisting the other. Concretely, it is a policy-gradient-assisted MAP-Elites archive, generalised toward the constrained actor-critic treatment of diversity introduced by quality-diversity actor-critic. A gradient-free evolutionary operator supplies global exploration and is indifferent to the non-smooth dynamics, discrete chiller staging and dew-point cliffs, that defeat gradient methods. A soft actor-critic operator, in turn, supplies sample-efficient, fine-grained modulation of continuous setpoints. \Cref{fig:algorithm} shows details of the algorithm.

\subsection{Algorithm}

\subsubsection{Contextual archive}

The archive is a product structure rather than a single tessellation. One behaviour-indexed sub-archive of twelve centroids is maintained per context cell, so a genome is scored inside one context and located there by behaviour:
\begin{equation}
\text{archive}[i][j], \qquad i \in \{1,\dots,N_c\}, \qquad j \in \{1,\dots,N_b\}
\label{eq:archive}
\end{equation}
Where $i$ indexes the context cell, $j$ indexes the behaviour niche within it. $N_c$, $N_b$ are the number of context centroids and behaviour centroids per context respectively.

Context and behaviour are assigned independently, each by nearest centroid in its own space:
\begin{equation}
c(\text{day}) = \arg\min_{c} \left\lVert \pi(F_{\text{day}}) - \mu_c \right\rVert, \qquad
j^*(\theta) = \arg\min_{j} \left\lVert \mathrm{bd}(\theta) - \nu_j \right\rVert
\label{eq:assignment}
\end{equation}
Here $\pi(\cdot)$ denotes the two-component principal-component projection of the daily feature vector $F_{\text{day}}$, and $\mu_c$ is the centroid of context cell $c$. The term $\mathrm{bd}(\theta)$ is genome $\theta$'s behaviour descriptor, and $\nu_j$ is the centroid of behaviour niche $j$.

The behaviour descriptor $\mathrm{bd}(\theta)$ is read directly from the policy rather than from a closed-loop trace. It is formed by feeding a fixed synthetic probe battery to the actor and summarising the emitted actions:
\begin{equation}
\mathrm{bd}(\theta) = (g_T,\, g_W,\, p_n,\, p_d) \in \mathcal{B} \subset \mathbb{R}^4
\label{eq:bd}
\end{equation}
Here $g_T$ and $g_W$ are the regression slopes of the chilled-water setpoint and tower-fan speed actions against outdoor temperature and humidity. The terms $p_n$ and $p_d$ are the fraction of unoccupied probe points requesting the plant on, split below and above a twenty-nine-degree drift threshold. Pooling the two plant-on fractions was tested and concealed an eight-fold difference in behaviour, which is why the split is kept.

Fitness is normalised against a precomputed Guideline 36 return for the same calendar day before insertion. This step collapses between-context return variance from a measured standard deviation to effectively zero:
\begin{equation}
\tilde{F}(\theta) = \frac{F(\theta,\text{day}) - F_{G36}(\text{day})}{F_{G36}(\text{day})}
\label{eq:fitness}
\end{equation}
where $F(\theta,\text{day})$ is the undiscounted return of genome $\theta$ over the one-day episode. $F_{G36}(\text{day})$ is the precomputed Guideline 36 return for that same calendar day.

A candidate is written into $\text{archive}[c(\text{day})][j^*(\theta)]$ only if that niche is empty or its stored normalised fitness is exceeded, and discarded otherwise. Search quality is reported by coverage, the fraction of the two hundred and sixteen niches filled, measured at 99.5\%. A quality-diversity score, the sum of positive elite fitness, is reported alongside the control key-performance indicators.

\subsubsection{Variation operator}

The archive is improved each generation by two operators in a fixed proportion. The evolutionary operator applies directed iso-line-directional variation over two random elites, which is reward-agnostic and embarrassingly parallel and escapes the local optima created by the non-smooth staging and dew-point dynamics:
\begin{equation}
\theta' = \theta_i + \sigma_1\, \mathcal{N}(0,I) + \sigma_2\, (\theta_j - \theta_i)\, \mathcal{N}(0,1)
\label{eq:variation}
\end{equation}
where $\theta_i$ and $\theta_j$ are two sampled elites, $\sigma_1$ the isotropic mutation scale, and $\sigma_2$ the directional scale.

The policy-gradient operator improves a batch of elites by Soft Actor-Critic (SAC) gradient steps that use twin critics trained on the shared replay buffer. SAC is chosen over on-policy alternatives because off-policy replay reuses the population rollouts and maximum-entropy exploration suits the coupled temperature-humidity action space.

The critic minimises the clipped double-Q entropy-regularised error:
\begin{equation}
J_Q(\psi_j) = \mathbb{E}_{(s,u,r,s')\sim\mathcal{D}}\left[\tfrac{1}{2}\left(Q_{\psi_j}(s,u) - y\right)^2\right], \qquad
y = r + \gamma \min_{j} Q_{\bar\psi_j}(s',u') - \alpha \log \pi(u'\mid s')
\label{eq:critic}
\end{equation}
with $Q_{\psi_j}$ the critic networks, $\mathcal{D}$ the replay buffer, $y$ the entropy-regularised target, $\bar\psi_j$ the target-network parameters, $u' \sim \pi$ the next sampled action, and $\alpha$ the temperature.

The actor, which is the policy-gradient variation on the genome, ascends the entropy-regularised value:
\begin{equation}
J_\pi(\theta) = \mathbb{E}_{s\sim\mathcal{D},\, u\sim\pi_\theta}\left[\alpha \log \pi_\theta(u\mid s) - \min_{j} Q_{\psi_j}(s,u)\right], \qquad
\theta \leftarrow \theta - \eta\, \nabla_\theta J_\pi(\theta)
\label{eq:actor}
\end{equation}
with $\eta$ the learning rate. The temperature is auto-tuned to a target entropy $\mathcal{H}$ by minimising:
\begin{equation}
J(\alpha) = \mathbb{E}_{u\sim\pi}\left[-\alpha \log \pi(u\mid s) + \mathcal{H}\right]
\label{eq:temperature}
\end{equation}

\subsubsection{Synergistic coupling}

Two exchanges make the scheme synergistic rather than two parallel searches. First, every fixed number of generations, the current policy-gradient actor is evaluated and inserted into the archive by the normal rule. A high-return region discovered by the gradient learner thereby seeds the corresponding niche, and is otherwise discarded. Second, as elite rollouts already populate the shared buffer, the critics learn from the diverse, wide-coverage data produced by evolution. A new actor may also be periodically warm-started by behaviour-cloning the current highest-fitness elite, which cures the cold-start problem.

Each generation produces offspring splitting evenly between the two variation operators. The evolutionary operator draws two archive elites and applies a directional isotropic-plus-line mutation that exploits the empirical correlation between high-performing genomes. The policy-gradient operator draws an elite and takes ascent steps against twin critics trained on a single replay buffer shared across every offspring's rollout, in the entropy-regularised form of soft actor-critic. Every defined number of generations the running actor itself, rather than an offspring, is rolled out and inserted, so the archive cannot drift far from what the shared critic currently favours.

\subsection{Implementation}

Context is defined empirically rather than assumed. Each calendar day is summarised by four features: mean cooling load, latent-load fraction, peak wet-bulb temperature and daily solar irradiation. Projected onto their top two principal components, these jointly explain 77.1\% of the feature variance. K-means clustering of the 363 non-extreme days into sixteen cells, plus two fixed sentinel centroids for the measured peak-load and minimum-solar days, gives eighteen context cells in total. This construction was itself a response to a measured failure of an earlier, context-blind archive design. Across ten near-converged policies evaluated on six representative days, 92.9\% of return variance was attributable to which day was drawn rather than to the genome under evaluation. A single trace-based behaviour axis, moreover, moved more than twenty times further across contexts than across policies. Both problems are addressed by making context an explicit archive axis and normalising fitness within it, which reduces the between-context standard deviation of normalised return to effectively zero.

Each context cell holds its own behaviour-indexed archive of twelve centroids, so the full archive is a product of eighteen contexts by twelve behaviour niches. The behaviour descriptor is computed by probing the actor with a fixed, synthetic series of observations spanning the measured annual range of zone temperature, humidity, outdoor conditions, occupancy and time of day, rather than from the trace of an actual episode. This guarantees the descriptor is a pure function of the policy parameters and provably invariant to whichever context the policy happens to be evaluated in. Its four axes are the sensitivity of the chilled-water setpoint to outdoor temperature, the sensitivity of tower-fan speed to outdoor humidity, and the fraction of unoccupied probe points at which the policy requests the plant on. This last axis is split by whether the zone is drifting warm or within its normal unoccupied band, because pooling the two regimes concealed an eight-fold difference in behaviour.

Training addresses partial observability of the thermal-mass state through a warm-start bank rather than a recurrent encoder. A full-year reference rollout under always-on Guideline 36 records the zone-air, thermal-mass and humidity state at the start of occupancy for every calendar day, stratified by day of week. This stratification is necessary because Monday's thermal mass measurably carries the preceding weekend's heat, while the zone-air temperature itself does not. Each training episode begins by drawing a context-representative day and a matching day-of-week thermal state from this bank, rather than resetting to an idealised isothermal condition. The resulting one-day rollout's return is then normalised against a precomputed Guideline 36 return for that same day before being written into the archive.

Every candidate action, in training and at deployment alike, is projected onto a feasible set by a deterministic safety shield enforcing relative-humidity and dew-point ceilings, actuator ranges, ramp limits and chiller minimum on-off times. This is the single most consequential mechanism in the system, since the measured relative-humidity-ceiling violation rate is zero for every controller evaluated, a hard projection rather than a reward penalty producing that guarantee directly.

At deployment, a daily forecast selects the nearest context cell, committed only after two consecutive confirming days to suppress boundary noise that would otherwise flip context at roughly 80\% of day boundaries. Within the active context, a windowed estimate of the same four behaviour statistics selects the nearest behaviour niche at every control step. If no niche is feasible, dispatch falls back first to the nearest populated niche in another context, and finally to a rule-based controller. This paper reports the architecture as implemented and evaluated, not as a proposal. Whether the evolutionary component earns its computational cost over a soft-actor-critic-only baseline, and whether the contextual archive currently beats Guideline 36 on energy as well as on safety, are treated as open questions. The case study that follows measures both rather than assuming either in the design.

\section{HVAC control optimization}
\label{sec:case-study}

The case study is a cooling-only, water-cooled commercial building representative of the Singapore stock, in which air-conditioning is the dominant end use and the central chilled-water plant dominates air-conditioning. The reference topology follows the BOPTEST ``multizone\_office\_complex\_air'' template, a 42,757-square-metre building collapsed to three representative floors with five zones per floor, four perimeter and one core, retaining water-cooled central-plant and variable-air-volume structure while removing the boiler, terminal reheat and dry-bulb economiser. The plant comprises several water-cooled centrifugal chillers sized against the annual peak load, cooling towers, primary and secondary chilled-water pumps and condenser-water pumps, feeding air-handling units with chilled-water cooling coils and variable-air-volume terminals. Three climate facts fix which sub-models matter. The wet-bulb temperature at Singapore-Changi sits near 25 to 26\textdegree C and varies little across the year. Consequently, the cooling tower's approach, rather than any economiser opportunity, is the binding constraint on condenser-water optimisation. The design sensible heat ratio is approximately 0.71, so dehumidification rather than temperature sizes the coil. Comfort and humidity limits, 23 to 25\textdegree C, a relative-humidity ceiling of 65\% and a Predicted-Mean-Vote band of plus or minus 0.5, follow SS 553 and are treated as safety constraints rather than reward terms.

The environment is formulated as a continuous-time differential-algebraic system, discretised at the control step and wrapped as a partially observed Markov decision process, because thermal-mass and moisture states are not directly measurable. Two coupled blocks are solved at different cadences. A zone and air-side block, carrying zone temperature, mass temperature, humidity ratio and coil outlet state, updates on a 5-minute physics sub-step. A plant and water-side block, carrying chilled-water and condenser-water temperatures and chiller staging, updates on a 20-minute control step. This decoupling costs only 0.09\% additional annual energy relative to matched 5-minute stepping, while quadrupling agent throughput. Each zone is represented by a lumped three-resistance, two-capacitance thermal network with an air node and a mass node. Zone moisture follows a well-mixed humidity-ratio balance driven by supply air, infiltration and internal generation, closed by the ASHRAE psychrometric relations for relative humidity and dew point. The cooling coil is represented by an apparatus-dew-point, bypass-factor model that couples sensible and latent duty and closes onto the chilled-water side. The chiller is represented by a DOE-2-style bi-quadratic performance-curve model, and the cooling tower by a Merkel effectiveness relation against the ambient wet bulb. Fan and pump power follow the cube-law affinity relation. Chiller staging introduces binary on-off variables with minimum-capacity and minimum-run-time constraints, so any optimisation-based controller over this environment is intrinsically a mixed-integer nonlinear program.

The action available to the supervisory controller is a four-dimensional vector, the chilled-water supply-temperature setpoint, the condenser-water and tower-fan speed, a plant-enable request and a chiller-stage request. Every requested action passes through a deterministic safety shield, box and ramp clipping followed by a humidity-triggered tightening of the chilled-water and fan limits, and then through hard overrides, a minimum on-off timer, a humidity and dew-point guard, an over-temperature guard and a comfort guard, any of which can force the plant on regardless of the policy's request. The reward is a three-term, dimensionless combination of integrated plant electrical energy, occupancy-weighted predicted-mean-vote discomfort outside a comfort deadband, and a shield-reliance penalty charging the displacement between the requested and the executed action. Comfort, humidity and equipment-protection limits are thereby enforced as hard constraints rather than tuned through reward weights, and the reward instead prices only what remains inside the feasible set.

Training a population-based, archive-filling search of this kind requires many millions of environment steps, infeasible against a high-fidelity equation-based twin. The environment is accordingly built in two tiers. A reduced-order tier, the differential-algebraic system above integrated with a semi-implicit scheme at a 1-minute sub-step, runs at microsecond-to-millisecond cost per step and carries the training rollouts. A calibrated, higher-fidelity Modelica or EnergyPlus twin carries periodic validation, accepted against the fidelity criteria of ASHRAE Guideline 14: a root-mean-square-error coefficient of variation at or below 30\% and a normalised mean bias at or below 10\%. It is reconciled against the reduced-order tier over the operating envelope the controller actually visits. This case study additionally uses the BOPTEST containerised benchmarking framework as the reference case and validation harness. This is a defensible choice: it lets the reduced-order identification pipeline and the controller be assessed on a reproducible, community-standard emulator against fixed key-performance indicators. It is explicitly not treated as a source of Singapore climate calibration, since no released BOPTEST case combines a water-cooled tower plant with humidity as a control objective.

The case study is evaluated on a full-year (8,760-hour) sequential backtest rather than isolated daily episodes. This way, overnight and weekend thermal-mass carryover, and the resulting Monday-morning recovery penalty, are represented rather than assumed away. Key-performance indicators including annual energy, chiller starts, peak zone temperature, relative-humidity violation rate and occupied comfort-violation rate, are aggregated with the same real-calendar weighting used to build the training context archive. ASHRAE Guideline 36 is used as the baseline anchoring the evaluation. This baseline implements the standard's own trim-and-respond duct-static and supply-air-temperature reset with dual-maximum terminal logic, run continuously. It is the incumbent every learned controller must beat, rather than a fixed-setpoint strawman.

\section{Results and discussion}
\label{sec:results}

\begin{figure}[!ht]
	\centering
	\includegraphics[width=\textwidth]{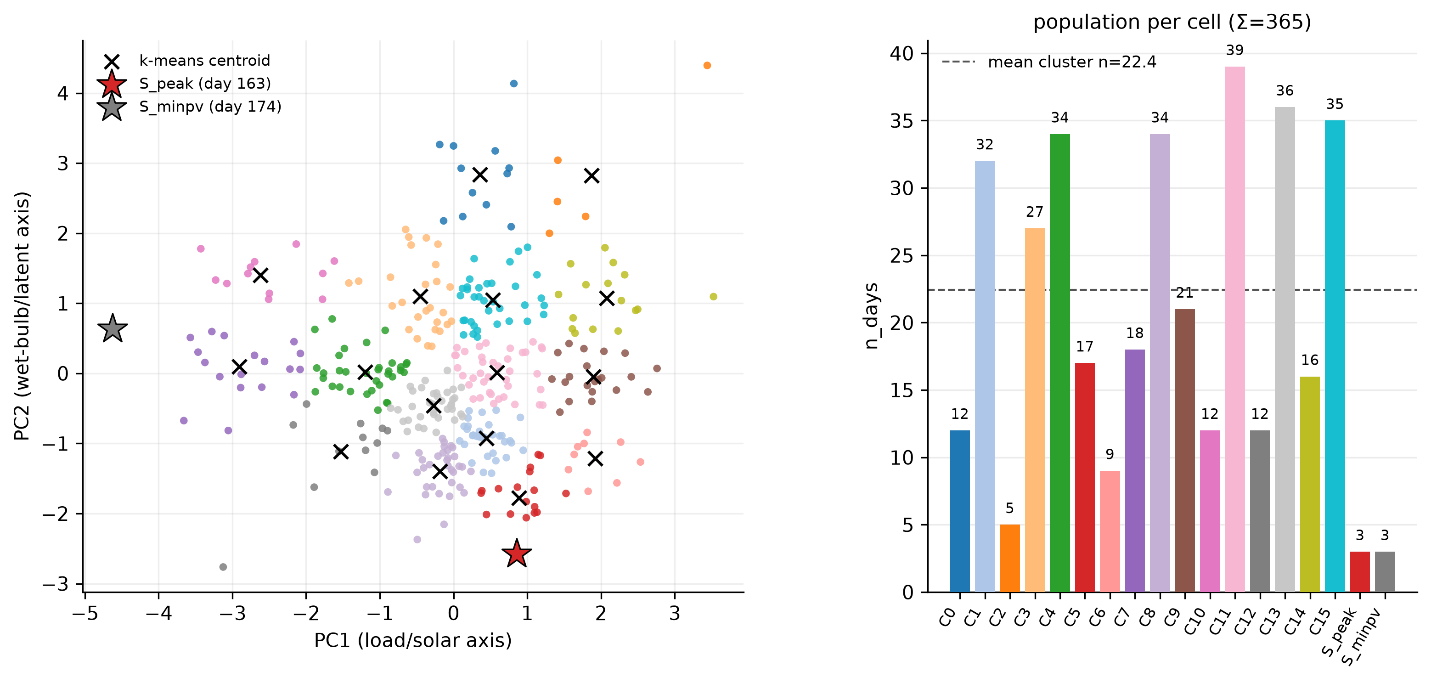}
	\caption{Context archive: principal-component basis and the cell map with day populations.}
	\label{fig:context-archive}
\end{figure}

The operating-context tessellation indexes the product archive on a daily vector of four features: mean cooling load, latent fraction, maximum wet-bulb temperature and daily global horizontal irradiance. Its principal-component basis was fitted on 363 non-extreme days, with the peak-cooling and minimum-photovoltaic days withheld as sentinels as shown in \Cref{fig:context-archive}. The first two components carry 43.6 and 33.5\% of variance (77.1\% jointly), supporting a two-dimensional tessellation. The first component is a load-and-solar axis, contrasting irradiance and mean load against latent fraction. The second is a near-orthogonal wet-bulb and latent-severity axis, loading on maximum wet bulb and mean load. Projecting all 365 days onto this plane and partitioning it into 16 k-means cells plus two fixed sentinel centroids yields an 18-cell archive. The k-means seed was chosen under a minimum-population constraint of five days per cluster; populations average 22.4 days, ranging from 5 to 39. The two sentinels, the peak-cooling and minimum-photovoltaic days, sit outside the main cloud and carry three days each, placed by construction.

\begin{figure}[!ht]
	\centering
	\includegraphics[width=\textwidth]{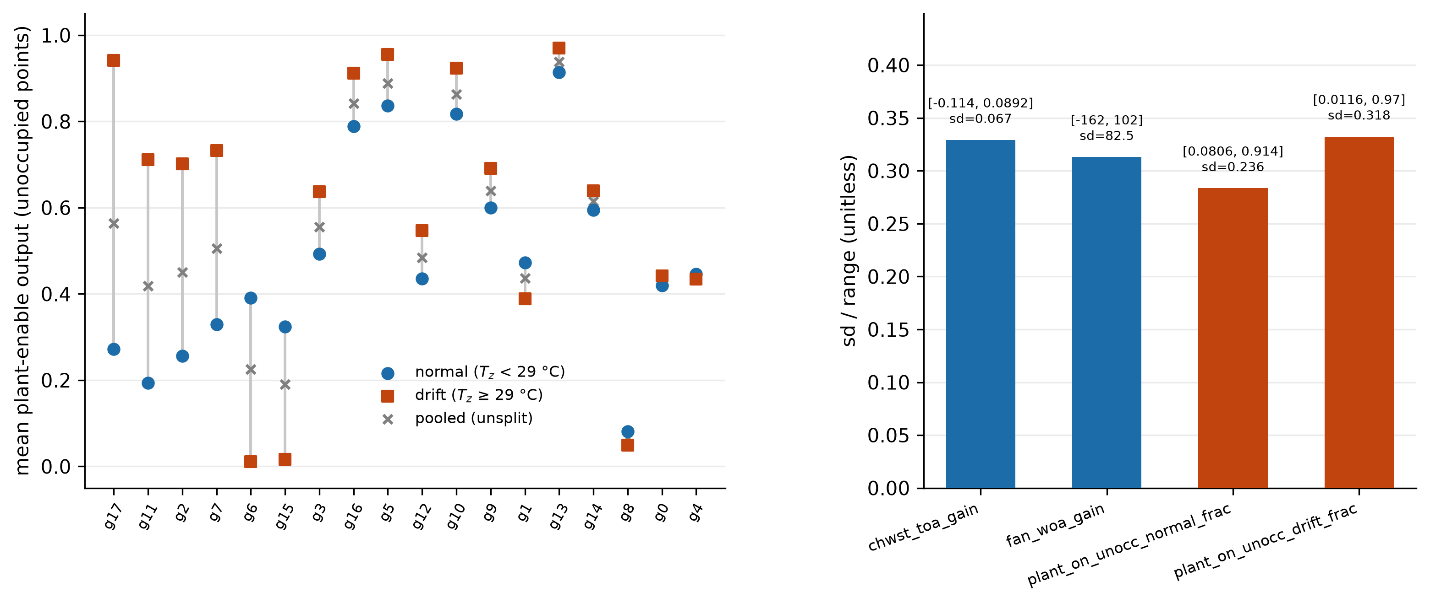}
	\caption{Policy-probe behaviour descriptor: (left) regime split vs.\ pooled (genomes sorted by gap), (right) axis discrimination of random genomes (sd normalised by each axis's own min-max span).}
	\label{fig:behavior-descriptor}
\end{figure}

\Cref{fig:behavior-descriptor} justifies the second archive axis, the policy-probe behaviour descriptor, obtained by feeding a fixed synthetic suite of observations to the actor and summarising emitted actions. Controller behaviour, not operating conditions, should distinguish genomes: with probe inputs held constant, the descriptor is a pure function of the genome. A context-invariance check confirms the substitution is exact: probing three genomes before and after a full episode rollout returns zero deviation on every axis. The probe grid sweeps seven variables, including zone temperature, humidity, occupancy and prior chilled-water setpoint, drawing 4,500 of 9,072 Cartesian points per evaluation, with wet-bulb derived psychrometrically for consistency. All four retained axes discriminate comparably once normalised by attainable span (dispersions 0.28--0.34 across genomes), despite raw scales differing by three orders of magnitude. Splitting unoccupied plant-enable behaviour into normal and drift regimes at 29\textdegree C recovers a 0.67 separation that pooling eliminates. The two gain axes, however, remain pooled: their occupied and unoccupied slopes correlate at 0.984, so splitting would waste a dimension.

Five independent seeds were trained for the CQD-ERL and SAC learners. Each was dispatched over an identical 365-day backtest at a 20-minute control step under the same weather, load and hard safety shield. The ASHRAE Guideline 36 sequence (G36) is seed-independent and was evaluated once. Dispersion is the sample standard deviation over five seeds; learner comparisons are two-sided Welch t-tests. Normalised fitness $\tilde{F}$ is referenced to G36, so $\tilde{F} > 0$ denotes improvement.

\begin{figure}[!ht]
	\centering
	\includegraphics[width=\textwidth]{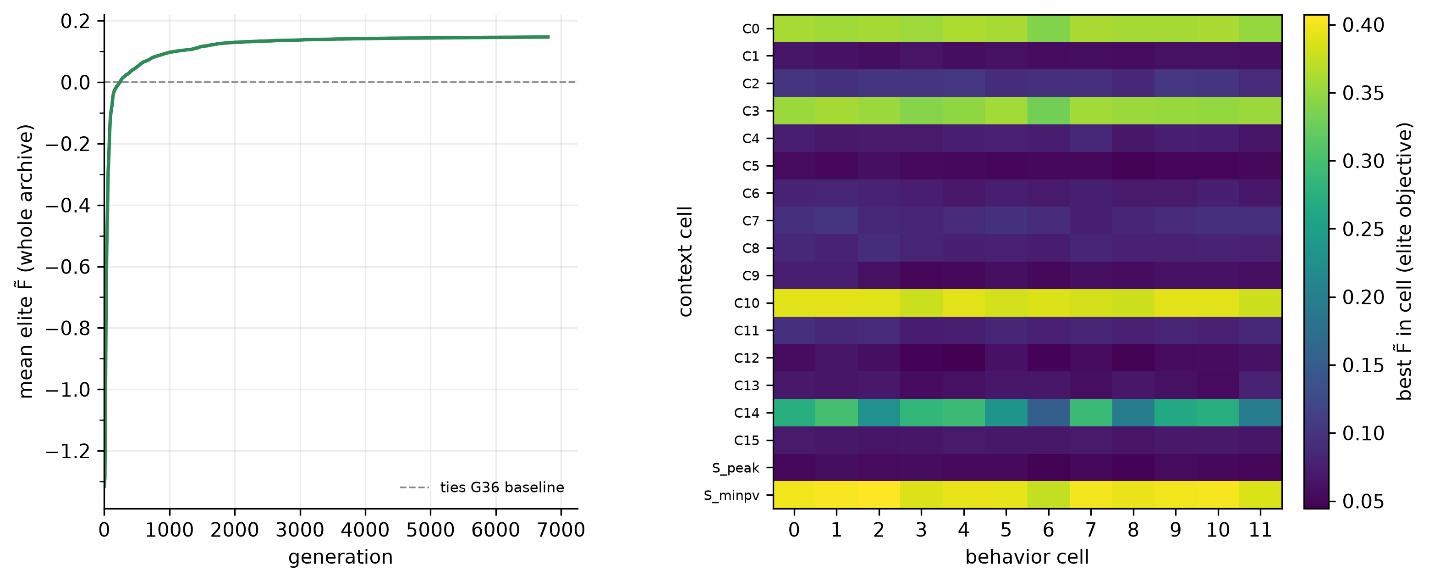}
	\caption{Coverage-normalised mean elite fitness in the training loop, and the elite-fitness map.}
	\label{fig:coverage-map}
\end{figure}

The results of the CQD-ERL arm show that the archive converges early, completely and reproducibly. The cell product archive filled completely in every seed. Coverage reached 50\% after 18,761 environment steps, 90\% after 56,371 $\pm$ 9,833, and unity after 126,889 $\pm$ 15,324. 95\% of the final quality-diversity score was attained after 91,630 $\pm$ 12,875 steps. Search is therefore coverage-saturating early and quality-improving slowly thereafter (\Cref{fig:coverage-map}, left). Reproducibility is unusual for a population method. The quality-diversity score converged to 2190.70 $\pm$ 0.36 (coefficient of variation 0.017\%) and coverage to 1.000 with zero dispersion. The best elite reached $\tilde{F} = 0.404 \pm 0.004$, the archive mean $0.1421 \pm 0.0017$. All 1,080 stored elites across five seeds satisfy $\tilde{F} > 0$, the weakest at 0.043. Every member of the deployed portfolio therefore outperforms G36 on its own context, so dispatch never has to protect the plant from its own archive.

The elite-fitness map (\Cref{fig:coverage-map}, right) shows strong row structure. A two-way variance decomposition confirms it: the context factor explains 99.43 $\pm$ 0.26\% of between-cell fitness variance, the behaviour factor 0.079 $\pm$ 0.044\%. This is the descriptor-selection criterion satisfied to a strict degree: the behaviour axis is nearly orthogonal to quality, so the twelve niches per context hold distinct strategies of near-equivalent merit rather than twelve rungs of a quality ladder.

Both learners reduce annual energy by a similar and statistically indistinguishable margin: 3.40 $\pm$ 0.43\% (95\% CI 2.86--3.94) for the archive and 3.67 $\pm$ 0.49\% (95\% CI 3.07--4.27) for the single SAC policy (Welch $p = 0.383$). On aggregate energy the evolutionary component buys nothing. It buys three other things: a 4.95$\times$ faster dispatch (1.29 ms versus 6.39 ms per control step, against 0.96 ms for G36 itself), part-load efficiency reported below, and reproducibility. The last is noticeable: annual chiller starts vary by 1.0\% across CQD-ERL seeds and 18.9\% across SAC seeds, a variance ratio of 272. The SAC learner delivered a plant that started 2,849 times in one seed and 3,388 in another: an 18.9\% spread in the quantity that sets compressor maintenance interval. The CQD-ERL does not produce a better expected outcome but a far more commissionable one from the same data.

Conditioning on evaporator load removes the scheduling confound. \Cref{tab:power} presents the resulting load-weighted specific power in kW/RT at different load bands. The CQD-ERL improves on G36 in every band. The largest margins lie in the 100--400 RT part-load range: 10.4 to 17.8\% at matched duty. This is the band where staging, chilled-water reset and condenser-water reset are simultaneously free, and where a reactive trim-and-respond sequence is structurally weakest. In the same band the SAC policy achieves only 5.7 to 10.7\%, so the archive is 5.5 and 11.9 percentage points better at 100--200 RT and 200--300 RT respectively. Aggregate parity between the two learners is explained by the 3,300--3,450 annual plant-on hours above 600 RT, where neither improves on the baseline. The converged strategy is legible and physically sensible. It runs chilled-water supply at 8.89\textdegree C against 7.68\textdegree C for G36 (+1.21 K, directly reducing compressor lift), with 47\% less setpoint modulation, 2.88 chillers staged against 2.08 (spreading duty across more lightly loaded machines), and 11\% lower tower-fan speed.

\begin{table}[t]
	\caption{Load-weighted specific power (kW/RT).}
	\centering
	\begin{tabular}{lrrrr}
		\toprule
		Load band (RT) & G36 h/yr & G36 (kW/RT) & CQD-ERL & SAC \\
		\midrule
		$<$ 100    & 4,343 & 1.413 & $-$3.0\%  & $-$25.6\% \\
		100--200   & 275   & 0.838 & $-$11.2\% & $-$5.7\%  \\
		200--300   & 205   & 0.843 & $-$17.8\% & $-$5.9\%  \\
		300--400   & 150   & 0.683 & $-$10.4\% & $-$10.7\% \\
		400--600   & 484   & 0.662 & $-$3.7\%  & $-$4.5\%  \\
		\bottomrule
	\end{tabular}
	\label{tab:power}
\end{table}

\Cref{tab:quartile} shows the calculated saving by outdoor wet-bulb quartile where the saving falls monotonically and reverses sign in the top quartile. As the wet bulb approaches its 28.17\textdegree C maximum, the tower approach margin closes and the condenser-water reset loses authority. The rising latent load, meanwhile, drives plant-on from 0.458 to 0.760 and removes the scheduling lever. In the quartile carrying 37\% of annual energy, both principal degrees of freedom are saturated. This constraint, previously argued on physical grounds, now appears as a measured quantity. It locates the residual opportunity in load shifting and storage rather than in better setpoints.

\begin{table}[t]
	\caption{Saving by outdoor wet-bulb quartile (edges 21.07, 24.44, 25.14, 25.79, 28.17\textdegree C).}
	\centering
	\begin{tabular}{lrrrr}
		\toprule
		Quartile & Share of annual energy & CQD-ERL & SAC & Plant-on \\
		\midrule
		Q1 ($<$ 24.44\textdegree C) & 14.7\% & $-$9.50\% & $-$11.68\% & 0.458 \\
		Q2                          & 21.3\% & $-$6.36\% & $-$6.32\%  & 0.526 \\
		Q3                          & 27.0\% & $-$2.97\% & $-$3.26\%  & 0.620 \\
		Q4 ($>$ 25.79\textdegree C) & 37.0\% & $+$0.41\% & $+$0.75\%  & 0.760 \\
		\bottomrule
	\end{tabular}
	\label{tab:quartile}
\end{table}

Across 5 seeds, the relative-humidity violation rate was exactly zero, the safety-shield correction magnitude was exactly zero. The fallback hierarchy was not entered: no context-empty fallback, no G36 fallback, 100\% primary-tier service. Occupied comfort was preserved (0.36 $\pm$ 0.32\% of steps outside the PMV deadband, $p = 0.208$ against the single policy). The dispatch layer behaved as designed showing that the portfolio is genuinely exercised. All five seeds made exactly 45 context switches with a median dwell of 5.5 days: regime persistence on a weekly timescale, not chattering. 83.3\% of context cells and all behaviour niches were used, 50.3 $\pm$ 0.023\% of the (context, behaviour) pairs were dispatched at least once. Dispatch entropy reached 74.7\% of uniform, and the most-used niche accounted for only 4.3--6.0\% of steps.

\begin{figure}[!ht]
	\centering
	\includegraphics[width=\textwidth]{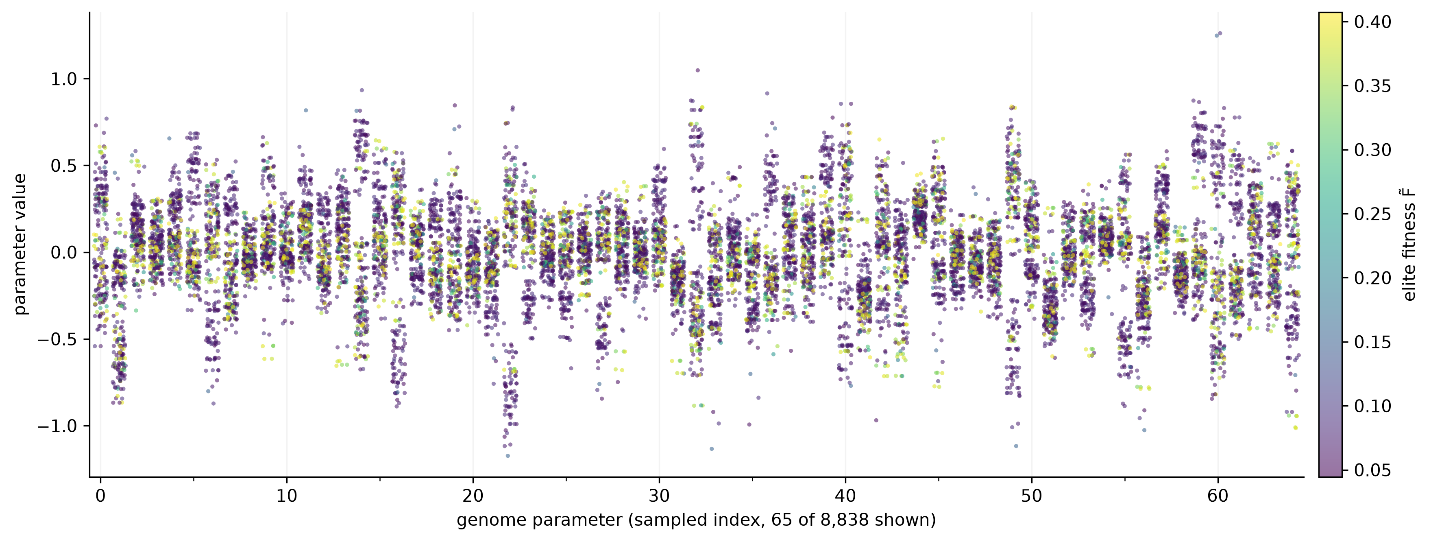}
	\caption{Elite-genome parameter concentration of the 216 elites and 8,838-dimensional genomes.}
	\label{fig:genome-pca}
\end{figure}

Policy-gradient offspring exceed iso-line-directional offspring in mean normalised fitness by 0.117, or 82\% of the mean elite fitness. The critics trained on the shared replay buffer therefore transfer usable per-step credit assignment into the population. Soft actor-critic actor injections, at 0.62\% of rollouts, carry the highest mean fitness of any source, confirming that archive-learner exchange is disproportionately valuable relative to its cost. The gradient-free operator retains the wider offspring distribution and is what sustains coverage at unity. The posited synergy is therefore observable in the training data rather than assumed. Principal-component analysis of the 216 stored 8,838-dimensional genomes reinforces the mechanism as shown in \Cref{fig:genome-pca}. The leading ten components capture a median 94.3\% of elite variance, confirming the elite-hypervolume structure the iso-line operator exploits.

The 3.40\% aggregate saving sits far below the 20--70\% of early deep reinforcement learning against unspecified rule-based baselines, and below the 11--15\% of field-validated model predictive control. Three reasons account for this and none is a deficiency of the controller. The baseline is Guideline 36 implemented as written. The correct comparators are the $\sim$17\% reported for low-level deep reinforcement learning against G36 in temperate climates with functioning economisers. The climate removes that lever entirely: air-side economising contributes nothing at a 25\textdegree C wet bulb. The reward is un-priced as the controller minimises kilowatt-hours under a flat tariff and cannot access tariff-driven savings. The contribution here is architectural before it is a savings number. This is the first archive-based quality-diversity evolutionary reinforcement-learning controller applied to a building energy system, dispatched over a full year with zero violations, zero shield corrections and zero fallbacks at rule-based inference cost.

\section{Conclusion}

This paper presented CQD-ERL, a contextual quality-diversity evolutionary reinforcement-learning controller for the water-cooled chiller plant and air side of a large commercial building in the humid tropics, with the trainable reduced-order environment. The controller maintains a product archive of elites indexed jointly by an operating-context tessellation and a behavioural tessellation over a policy-probe descriptor that is a pure function of the genome. The archive is filled by a gradient-free iso-line-directional operator and a soft actor-critic policy-gradient operator in equal proportion, sharing one replay buffer and one pair of twin critics. Every action is also projected onto the feasible set by a deterministic safety shield. At deployment there is no online learning and no online optimisation: the regime is projected into context space and the operating point into behaviour space, and the corresponding pre-validated elite acts.

Over five seeds and a full-year backtest, the controller reduced whole-plant energy by 3.40 $\pm$ 0.43\% against an ASHRAE Guideline 36 sequence, with zero relative-humidity violations, zero shield corrections and zero fallback activations across 131,405 dispatched control steps. Archive coverage reached unity in every seed after 2.1\% of the training budget, and all 1,080 elites outperformed the baseline on their own context. At matched evaporator load, the archive was 10.4 to 17.8\% more efficient than Guideline 36 in the 100--400 RT part-load band and 5.5 to 11.9 points better than an identically trained soft actor-critic policy. Besides, it was indistinguishable on annual energy but 4.95 times faster at inference (1.29 ms per control step) and far more reproducible (272:1 variance ratio in annual chiller starts).

The context tessellation uses four physical day-ahead features, but the framework is not restricted to physical quantities. Incorporating electricity tariff structure would enable distinct on-peak and off-peak strategies, requiring reward re-pricing and reactivation of an existing observation channel. This is the most promising route to further cost reduction, since 37\% of annual energy falls where efficiency gains are unavailable. Demand response, the scenario the architecture was designed for, remains untested. An event-signal context axis would let dispatch switch to a pre-validated curtailment elite within one control step, with the shield guaranteeing the humidity ceiling is not breached. Marginal carbon intensity, locational price, thermal and electrical storage, and measured occupancy admit similar treatment. Remaining priorities include completing two-tier validation, separating the two components of the saving, repeating the transplant test with maximum-order correction, and validating the confirmation rule under forecast error. A field trial beginning with hardware-in-the-loop evaluation and a supervised advisory-mode trial is the natural next step.

\bibliographystyle{unsrtnat}
\bibliography{lib}

\end{document}